\documentclass[11pt,a4paper]{article}
\usepackage[hyperref]{acl2019}
\usepackage{times}
\usepackage{latexsym}
\usepackage{graphicx}
\usepackage{amsmath}

\aclfinalcopy 

\title{Energy and Policy Considerations for Deep Learning in NLP}

\author{Emma Strubell \qquad Ananya Ganesh \qquad Andrew McCallum\\
  College of Information and Computer Sciences \\
  University of Massachusetts Amherst \\
  {\tt \{strubell, aganesh, mccallum\}@cs.umass.edu} \\}

\date{}

\begin{document}
\maketitle
\begin{abstract}

  
  Recent progress in hardware and methodology for training neural networks has ushered in a new generation of large networks trained on abundant data. These models have obtained notable gains in accuracy across many NLP tasks. However, these accuracy improvements depend on the availability of exceptionally large computational resources that necessitate similarly substantial energy consumption. As a result these models are costly to train and develop, both financially, due to the cost of hardware and electricity or cloud compute time, and environmentally, due to the carbon footprint required to fuel modern tensor processing hardware. In this paper we bring this issue to the attention of NLP researchers by quantifying the approximate financial and environmental costs of training a variety of recently successful neural network models for NLP. Based on these findings, we propose actionable recommendations to reduce costs and improve equity in NLP research and practice. 

\end{abstract}

\section{Introduction}

  Advances in techniques and hardware for training deep neural networks have recently enabled impressive accuracy improvements across many fundamental NLP tasks \citep{Bahdanau2014, D15-1166,dozat2017deep,vaswani2017attention}, with the most computationally-hungry models obtaining the highest scores  \citep{peters2018deep,devlin2018bert,radford2019language, so2019evolved}. As a result, training a state-of-the-art model now requires substantial computational resources which demand considerable energy, along with the associated financial and environmental costs. Research and development of new models multiplies these costs by thousands of times by requiring re-training to experiment with model architectures and hyperparameters.
  Whereas a decade ago most NLP models could be trained and developed on a commodity laptop or server, many now require multiple instances of specialized hardware such as GPUs or TPUs, therefore limiting access to these highly accurate models on the basis of finances.
  
  \begin{table}[t!]
    \centering
    \begin{tabular}{lrr}
        \bf Consumption  & \bf CO$_\mathbf{2}$e (lbs)\\ \hline
        Air travel, 1 passenger, NY$\leftrightarrow$SF & 1984\\
        Human life, avg, 1 year & 11,023\\
        American life, avg, 1 year & 36,156 \\
        Car, avg incl. fuel, 1 lifetime & 126,000\\
        & & \\
        \bf Training one model (GPU)  & \\ \hline
        NLP pipeline (parsing, SRL) & 39 \\
        \ \ \ \ w/ tuning \& experimentation  & 78,468 \\
        Transformer (big)  & 192\\ 
        \ \ \ \ w/ neural architecture search & 626,155\\
    \end{tabular}
    \caption{Estimated CO$_2$ emissions from training common NLP models, compared to familiar consumption.\footnotemark \label{tab:footprints}}
\end{table}
\footnotetext{Sources: (1) Air travel and per-capita consumption: \protect\url{https://bit.ly/2Hw0xWc}; (2) car lifetime: \protect\url{https://bit.ly/2Qbr0w1}.}
  
  Even when these expensive computational resources are available, model training also incurs a substantial cost to the environment due to the energy required to power this hardware for weeks or months at a time. Though some of this energy may come from renewable or carbon credit-offset resources, the high energy demands of these models are still a concern since (1) energy is not currently derived from carbon-neural sources in many locations, and (2) when renewable energy is available, it is still limited to the equipment we have to produce and store it, and energy spent training a neural network might better be allocated to heating a family's home. It is estimated that we must cut carbon emissions by half over the next decade to deter escalating rates of natural disaster, and based on the estimated CO$_2$ emissions listed in Table~\ref{tab:footprints}, model training and development likely make up a substantial portion of the greenhouse gas emissions attributed to many NLP researchers.
  
  To heighten the awareness of the NLP community to this issue and promote mindful practice and policy, we characterize the dollar cost and carbon emissions that result from training the neural networks at the core of many state-of-the-art NLP models. We do this by estimating the kilowatts of energy required to train a variety of popular off-the-shelf NLP models, which can be converted to approximate carbon emissions and electricity costs. To estimate the even greater resources required to transfer an existing model to a new task or develop new models, we perform a case study of the full computational resources required for the development and tuning of a recent state-of-the-art NLP pipeline \citep{emnlp18-strubell}. We conclude with recommendations to the community based on our findings, namely: (1) Time to retrain and sensitivity to hyperparameters should be reported for NLP machine learning models; (2) academic researchers need equitable access to computational resources; and (3) researchers should prioritize developing efficient models and hardware.

\section{Methods}

To quantify the computational and environmental cost of training deep neural network models for NLP, we perform an analysis of the energy required to train a variety of popular off-the-shelf NLP models, as well as a case study of the complete sum of resources required to develop LISA \citep{emnlp18-strubell}, a state-of-the-art NLP model from EMNLP 2018, including all tuning and experimentation.

We measure energy use as follows. We train the models described in \S\ref{sec:models} using the default settings provided, and sample GPU and CPU power consumption during training. Each model was trained for a maximum of 1 day. We train all models on a single NVIDIA Titan X GPU, with the exception of ELMo which was trained on 3 NVIDIA GTX 1080 Ti GPUs. While training, we repeatedly query the NVIDIA System Management Interface\footnote{\texttt{nvidia-smi}: \protect\url{https://bit.ly/30sGEbi}} to sample the GPU power consumption and report the average over all samples. To sample CPU power consumption, we use Intel's Running Average Power Limit interface.\footnote{RAPL power meter: \protect\url{https://bit.ly/2LObQhV}}

\begin{table}[t]
\begin{center}
    \begin{tabular}{lrrrr}
         Consumer            & Renew.    & Gas   & Coal & Nuc. \\ \hline \hline
         China               & 22\%      &  3\%  & 65\% & 4\% \\
         Germany             & 40\%      & 7\%  & 38\% & 13\% \\ 
         United States       & 17\%      & 35\%  & 27\% & 19\% \\\hline
         Amazon-AWS          & 17\%      & 24\%  & 30\% & 26\% \\
         Google              & 56\%      & 14\%  & 15\% & 10\% \\
         Microsoft           & 32\%      & 23\%  & 31\% & 10\% \\
    \end{tabular}
\end{center}
\caption{Percent energy sourced from: Renewable (e.g. hydro, solar, wind), natural gas, coal and nuclear for the top 3 cloud compute providers \citep{clicking-clean}, compared to the United States,\footnotemark\ China\footnotemark\ and Germany \cite{burger2019net}. \label{tab:energy-breakdown}}
\end{table}
\footnotetext{U.S. Dept. of Energy: \url{https://bit.ly/2JTbGnI}}
\footnotetext{China Electricity Council; trans. China Energy Portal: \url{https://bit.ly/2QHE5O3}}

We estimate the total time expected for models to train to completion using training times and hardware reported in the original papers. We then calculate the power consumption in kilowatt-hours (kWh) as follows. Let $p_c$ be the average power draw (in watts) from all CPU sockets during training, let $p_r$ be the average power draw from all DRAM (main memory) sockets, let $p_g$ be the average power draw of a GPU during training, and let $g$ be the number of GPUs used to train. We estimate total power consumption as combined GPU, CPU and DRAM consumption, then multiply this by Power Usage Effectiveness (PUE), which accounts for the additional energy required to support the compute infrastructure (mainly cooling). We use a PUE coefficient of 1.58, the 2018 global average for data centers
\citep{uptime-datacenter}. Then the total power $p_t$ required at a given instance during training is given by:
\begin{align}
p_t = \frac{1.58t(p_c + p_r + gp_g)}{1000}
\end{align}
The U.S. Environmental Protection Agency (EPA) provides average CO$_2$ produced (in pounds per kilowatt-hour) for power consumed in the U.S. \citep{epa-egrid}, which we use to convert power to estimated CO$_2$ emissions:
\begin{align}
\mathrm{CO_2e} = 0.954 p_t
\end{align}
This conversion takes into account the relative proportions of different energy sources (primarily natural gas, coal, nuclear and renewable) consumed to produce energy in the United States. Table~\ref{tab:energy-breakdown} lists the relative energy sources for China, Germany and the United States compared to the top three cloud service providers. The U.S. breakdown of energy is comparable to that of the most popular cloud compute service, Amazon Web Services, so we believe this conversion to provide a reasonable estimate of CO$_2$ emissions per kilowatt hour of compute energy used.

\subsection{Models \label{sec:models}}

We analyze four models, the computational requirements of which we describe below. All models have code freely available online, which we used out-of-the-box. For more details on the models themselves, please refer to the original papers.

\noindent\textbf{Transformer}. 
The Transformer model \citep{vaswani2017attention} is an encoder-decoder architecture primarily recognized for efficient and accurate machine translation. The encoder and decoder each consist of 6 stacked layers of multi-head self-attention.
\citet{vaswani2017attention} report that the Transformer {\bf base} model (65M parameters) was trained on 8 NVIDIA P100 GPUs for 12 hours, and the Transformer {\bf big} model (213M parameters) was trained for 3.5 days (84 hours; 300k steps). This model is also the basis for recent work on neural architecture search ({\bf NAS}) for machine translation and language modeling \citep{so2019evolved}, and the NLP pipeline that we study in more detail in \S\ref{sec:case} \citep{emnlp18-strubell}. \citet{so2019evolved} report that their full architecture search ran for a total of 979M training steps, and that their base model requires 10 hours to train for 300k steps on one TPUv2 core. This equates to 32,623 hours of TPU or 274,120 hours on 8 P100 GPUs.

\noindent\textbf{ELMo.} 
The ELMo model \citep{peters2018deep} is based on stacked LSTMs and provides rich word representations in context by pre-training on a large amount of data using a language modeling objective. Replacing context-independent pre-trained word embeddings with ELMo has been shown to increase performance on downstream tasks such as named entity recognition, semantic role labeling, and coreference.
\citet{peters2018deep} report that ELMo was trained on 3 NVIDIA GTX 1080 GPUs for 2 weeks (336 hours).

\noindent\textbf{BERT.} 
The BERT model \citep{devlin2018bert} provides a Transformer-based architecture for building contextual representations similar to ELMo, but trained with a different language modeling objective. BERT substantially improves accuracy on tasks requiring sentence-level representations such as question answering and natural language inference. 
\citet{devlin2018bert} report that the BERT base model (110M parameters) was trained on 16 TPU chips for 4 days (96 hours). NVIDIA reports that they can train a BERT model in 3.3 days (79.2 hours) using 4 DGX-2H servers, totaling 64 Tesla V100 GPUs \citep{nvidia-bert}. 

\noindent\textbf{GPT-2}. 
This model is the latest edition of OpenAI's GPT general-purpose token encoder, also based on Transformer-style self-attention and trained with a language modeling objective \citep{radford2019language}. By training a very large model on massive data, \citet{radford2019language} show high zero-shot performance on question answering and language modeling benchmarks. The large model described in \citet{radford2019language} has 1542M parameters and is reported to require 1 week (168 hours) of training on 32 TPUv3 chips.
\footnote{Via the authors \href{https://bit.ly/2Tw1x4L}{on Reddit}.}


\begin{table*}[t]
\begin{center}
    \begin{tabular}{llrrrrl}
         Model & Hardware & Power (W) & Hours & kWh$\cdot$PUE & CO$_2$e & Cloud compute cost \\ \hline \hline
         Transformer$_{base}$ & P100x8 & 1415.78 & 12 & 27 & 26 & \$41--\$140 \\
         Transformer$_{big}$ & P100x8 & 1515.43 & 84 & 201 & 192 & \$289--\$981 \\
         ELMo & P100x3 & 517.66 & 336 & 275 & 262 & \$433--\$1472 \\
         BERT$_{base}$ & V100x64 & 12,041.51 & 79 & 1507 & 1438 & \$3751--\$12,571 \\
         BERT$_{base}$ & TPUv2x16 & --- & 96 & --- & --- & \$2074--\$6912 \\
         NAS & P100x8 & 1515.43 & 274,120 & 656,347 & 626,155 & \$942,973--\$3,201,722 \\
         NAS & TPUv2x1 & --- & 32,623 & --- & --- & \$44,055--\$146,848 \\
         GPT-2 & TPUv3x32 & --- & 168 & --- & --- & \$12,902--\$43,008 \\
    \end{tabular}
\end{center}
\caption{Estimated cost of training a model in terms of CO$_2$ emissions (lbs) and cloud compute cost (USD).\footnotemark\ Power and carbon footprint are omitted for TPUs due to lack of public information on power draw for this hardware.\label{tab:training}}
\end{table*}
\footnotetext{GPU lower bound computed using pre-emptible P100/V100 U.S. resources priced at \$0.43--\$0.74/hr, upper bound uses on-demand U.S. resources priced at \$1.46--\$2.48/hr. We similarly use pre-emptible (\$1.46/hr--\$2.40/hr) and on-demand (\$4.50/hr--\$8/hr) pricing as lower and upper bounds for TPU v2/3; cheaper bulk contracts are available.} 

\section{Related work}
There is some precedent for work characterizing the computational requirements of training and inference in modern neural network architectures in the computer vision community. 
\citet{Li2016EvaluatingTE} present a detailed study of the energy use required for training and inference in popular convolutional models for image classification in computer vision, including fine-grained analysis comparing different neural network layer types.
\citet{canziani2016analysis} assess image classification model accuracy as a function of model size and gigaflops required during inference. They also measure average power draw required during inference on GPUs as a function of batch size.
Neither work analyzes the recurrent and self-attention models that have become commonplace in NLP, nor do they extrapolate power to estimates of carbon and dollar cost of training.

Analysis of hyperparameter tuning has been performed in the context of improved algorithms for hyperparameter search \citep{bergstra2011algorithms,bergstra2012random,snoek2012practical}. 
To our knowledge there exists to date no analysis of the computation required for R\&D and hyperparameter tuning of neural network models in NLP.

\section{Experimental results\label{sec:experiments}}


\subsection{Cost of training \label{sec:training}}

Table~\ref{tab:training} lists CO$_2$ emissions and estimated cost of training the models described in \S\ref{sec:models}. Of note is that TPUs are more cost-efficient than GPUs on workloads that make sense for that hardware (e.g. BERT). We also see that models emit substantial carbon emissions; training BERT on GPU is roughly equivalent to a trans-American flight. \citet{so2019evolved} report that NAS achieves a new state-of-the-art BLEU score of 29.7 for English to German machine translation, an increase of just 0.1 BLEU at the cost of  at least \$150k in on-demand compute time and non-trivial carbon emissions.



\subsection{Cost of development: Case study \label{sec:case}}
To quantify the computational requirements of R\&D for a new model we study the logs of all training required to develop Linguistically-Informed Self-Attention \citep{emnlp18-strubell}, a multi-task model that performs part-of-speech tagging, labeled dependency parsing, predicate detection and semantic role labeling. This model makes for an interesting case study as a representative NLP pipeline and as a Best Long Paper at EMNLP.

Model training associated with the project spanned a period of 172 days (approx. 6 months). During that time 123 small hyperparameter grid searches 
were performed, resulting in 4789 jobs in total. Jobs varied in length ranging from a minimum of 3 minutes, indicating a crash, to a maximum of 9 days, with an average job length of 52 hours. All training was done on a combination of NVIDIA Titan X (72\%) and M40 (28\%) GPUs.\footnote{We approximate cloud compute cost using P100 pricing.} 

\begin{table}[h]
    \centering
    \begin{tabular}{llll}
        & & \multicolumn{2}{c}{Estimated cost (USD)} \\ \cline{3-4}
        Models & Hours & Cloud compute & Electricity \\ \hline \hline
        1 & 120 & \$52--\$175 & \$5 \\
        24 & 2880 & \$1238--\$4205 & \$118 \\
        4789 & 239,942 & \$103k--\$350k & \$9870

    \end{tabular}
    \caption{Estimated cost in terms of cloud compute and electricity for training: (1) a single model (2) a single tune and (3) all models trained during R\&D. \label{tab:case}}
\end{table}

The sum GPU time required for the project totaled 9998 days (27 years). This averages to about 60 GPUs running constantly throughout the 6 month duration of the project. Table~\ref{tab:case} lists upper and lower bounds of the estimated cost in terms of Google Cloud compute and raw electricity required to develop and deploy this model.\footnote{Based on average U.S cost of electricity of \$0.12/kWh.}
We see that while training a single model is relatively inexpensive, the cost of tuning a model for a new dataset, which we estimate here to require 24 jobs, or performing the full R\&D required to develop this model, quickly becomes extremely expensive.



\section{Conclusions}

\subsection*{Authors should report training time and sensitivity to hyperparameters.}
Our experiments 
suggest that it would be beneficial to directly compare different models to perform a cost-benefit (accuracy) analysis. To address this, when proposing a model that is meant to be re-trained for downstream use, such as re-training on a new domain or fine-tuning on a new task, authors should report training time and computational resources required, as well as model sensitivity to hyperparameters. This will enable direct comparison across models, allowing subsequent consumers of these models to accurately assess whether the required computational resources are compatible with their setting. More explicit characterization of tuning time could also reveal inconsistencies in time spent tuning baseline models compared to proposed contributions. Realizing this will require: (1) a standard, hardware-independent measurement of training time, such as gigaflops required to convergence, and (2) a standard measurement of model sensitivity to data and hyperparameters, such as variance with respect to hyperparameters searched.


\subsection*{Academic researchers need equitable access to computation resources.}
Recent advances in available compute come at a high price not attainable to all who desire access. Most of the models studied in this paper were developed outside academia; recent improvements in state-of-the-art accuracy are possible thanks to industry access to large-scale compute.

Limiting this style of research to industry labs hurts the NLP research community in many ways. First, it stifles creativity. Researchers with good ideas but without access to large-scale compute will simply not be able to execute their ideas, instead constrained to focus on different problems. Second, it prohibits certain types of research on the basis of access to financial resources. This even more deeply promotes the already problematic ``rich get richer'' cycle of research funding, where groups that are already successful and thus well-funded tend to receive more funding due to their existing accomplishments. Third, the prohibitive start-up cost of building in-house resources forces resource-poor groups to rely on cloud compute services such as AWS, Google Cloud and Microsoft Azure. 

While these services provide valuable, flexible, and often relatively environmentally friendly compute resources, it is more cost effective for academic researchers, who often work for non-profit educational institutions and whose research is funded by government entities, to pool resources to build shared compute centers at the level of funding agencies, such as the U.S. National Science Foundation.
For example, an off-the-shelf GPU server containing 8 NVIDIA 1080 Ti GPUs and supporting hardware can be purchased for approximately \$20,000 USD. At that cost, the hardware required to develop the model in our case study (approximately 58 GPUs for 172 days) would cost \$145,000 USD plus electricity, about half the estimated cost to use on-demand cloud GPUs. Unlike money spent on cloud compute, however, that invested in centralized resources would continue to pay off as resources are shared across many projects. A government-funded academic compute cloud would provide equitable access to all researchers. 

\subsection*{Researchers should prioritize computationally efficient hardware and algorithms.}
We recommend a concerted effort by industry and academia to promote research of more computationally efficient algorithms, as well as hardware that requires less energy. An effort can also be made in terms of software. There is already a precedent for NLP software packages prioritizing efficient models.
An additional avenue through which NLP and machine learning software developers could aid in reducing the energy associated with model tuning is by providing easy-to-use APIs implementing more efficient alternatives to brute-force grid search for hyperparameter tuning, e.g. random or Bayesian hyperparameter search techniques \cite{bergstra2011algorithms,bergstra2012random,snoek2012practical}. While software packages implementing these techniques do exist,\footnote{For example, the \href{https://github.com/hyperopt/hyperopt}{Hyperopt Python library}.} they are rarely employed in practice for tuning NLP models. This is likely because their interoperability with popular deep learning frameworks such as PyTorch and TensorFlow is not optimized, i.e. there are not simple examples of how to tune TensorFlow Estimators using Bayesian search. Integrating these tools into the workflows with which NLP researchers and practitioners are already familiar could have notable impact on the cost of developing and tuning in NLP.

\section*{Acknowledgements}
We are grateful to Sherief Farouk and the anonymous reviewers for helpful feedback on earlier drafts. This work was supported in part by the Centers for Data Science and Intelligent Information Retrieval, the Chan Zuckerberg Initiative under the Scientific Knowledge Base Construction project, the IBM Cognitive Horizons Network agreement no. W1668553, and National Science Foundation grant no. IIS-1514053. Any opinions, findings and conclusions or recommendations expressed in this material are those of the authors and do not necessarily reflect those of the sponsor.

\bibliography{acl2019}

\begin{thebibliography}{19}
\expandafter\ifx\csname natexlab\endcsname\relax\def\natexlab#1{#1}\fi

\bibitem[{Ascierto(2018)}]{uptime-datacenter}
Rhonda Ascierto. 2018.
\newblock \href
  {https://uptimeinstitute.com/2018-data-center-industry-survey-results}
  {{Uptime Institute Global Data Center Survey}}.
\newblock Technical report, Uptime Institute.

\bibitem[{Bahdanau et~al.(2015)Bahdanau, Cho, and Bengio}]{Bahdanau2014}
Dzmitry Bahdanau, Kyunghyun Cho, and Yoshua Bengio. 2015.
\newblock {Neural Machine Translation by Jointly Learning to Align and
  Translate}.
\newblock In \emph{3rd International Conference for Learning Representations
  (ICLR)}, San Diego, California, USA.

\bibitem[{Bergstra and Bengio(2012)}]{bergstra2012random}
James Bergstra and Yoshua Bengio. 2012.
\newblock Random search for hyper-parameter optimization.
\newblock \emph{Journal of Machine Learning Research}, 13(Feb):281--305.

\bibitem[{Bergstra et~al.(2011)Bergstra, Bardenet, Bengio, and
  K{\'e}gl}]{bergstra2011algorithms}
James~S Bergstra, R{\'e}mi Bardenet, Yoshua Bengio, and Bal{\'a}zs K{\'e}gl.
  2011.
\newblock Algorithms for hyper-parameter optimization.
\newblock In \emph{Advances in neural information processing systems}, pages
  2546--2554.

\bibitem[{Burger(2019)}]{burger2019net}
Bruno Burger. 2019.
\newblock \href {https://bit.ly/2EMRGyI} {{Net Public Electricity Generation in
  Germany in 2018}}.
\newblock Technical report, Fraunhofer Institute for Solar Energy Systems ISE.

\bibitem[{Canziani et~al.(2016)Canziani, Paszke, and
  Culurciello}]{canziani2016analysis}
Alfredo Canziani, Adam Paszke, and Eugenio Culurciello. 2016.
\newblock \href {http://arxiv.org/abs/1605.07678} {An analysis of deep neural
  network models for practical applications}.

\bibitem[{Cook et~al.(2017)Cook, Lee, Tsai, Kongn, Deans, Johnson, Jardim, and
  Johnson}]{clicking-clean}
Gary Cook, Jude Lee, Tamina Tsai, Ada Kongn, John Deans, Brian Johnson,
  Elizabeth Jardim, and Brian Johnson. 2017.
\newblock \href
  {https://storage.googleapis.com/planet4-international-stateless/2017/01/35f0ac1a-clickclean2016-hires.pdf}
  {{Clicking Clean: Who is winning the race to build a green internet?}}
\newblock Technical report, Greenpeace.

\bibitem[{Devlin et~al.(2019)Devlin, Chang, Lee, and
  Toutanova}]{devlin2018bert}
Jacob Devlin, Ming-Wei Chang, Kenton Lee, and Kristina Toutanova. 2019.
\newblock {BERT: Pre-training of Deep Bidirectional Transformers for Language
  Understanding}.
\newblock In \emph{NAACL}.

\bibitem[{Dozat and Manning(2017)}]{dozat2017deep}
Timothy Dozat and Christopher~D. Manning. 2017.
\newblock Deep biaffine attention for neural dependency parsing.
\newblock In \emph{ICLR}.

\bibitem[{EPA(2018)}]{epa-egrid}
EPA. 2018.
\newblock \href
  {https://www.epa.gov/energy/emissions-generation-resource-integrated-database-egrid}
  {{Emissions \& Generation Resource Integrated Database (eGRID)}}.
\newblock Technical report, U.S. Environmental Protection Agency.

\bibitem[{Forster et~al.(2019)Forster, Johnsen, Mandava, Sreenivas, Fu,
  Bernauer, Gray, Chetlur, and Puri}]{nvidia-bert}
Christopher Forster, Thor Johnsen, Swetha Mandava, Sharath~Turuvekere
  Sreenivas, Deyu Fu, Julie Bernauer, Allison Gray, Sharan Chetlur, and Raul
  Puri. 2019.
\newblock \href
  {https://web.archive.org/web/20190521104957/https://medium.com/future-vision/bert-meets-gpus-403d3fbed848}
  {{BERT Meets GPUs}}.
\newblock Technical report, NVIDIA AI.

\bibitem[{Li et~al.(2016)Li, Chen, Becchi, and Zong}]{Li2016EvaluatingTE}
Da~Li, Xinbo Chen, Michela Becchi, and Ziliang Zong. 2016.
\newblock Evaluating the energy efficiency of deep convolutional neural
  networks on cpus and gpus.
\newblock \emph{2016 IEEE International Conferences on Big Data and Cloud
  Computing (BDCloud), Social Computing and Networking (SocialCom), Sustainable
  Computing and Communications (SustainCom) (BDCloud-SocialCom-SustainCom)},
  pages 477--484.

\bibitem[{Luong et~al.(2015)Luong, Pham, and Manning}]{D15-1166}
Thang Luong, Hieu Pham, and Christopher~D. Manning. 2015.
\newblock \href {https://doi.org/10.18653/v1/D15-1166} {Effective approaches to
  attention-based neural machine translation}.
\newblock In \emph{Proceedings of the 2015 Conference on Empirical Methods in
  Natural Language Processing}, pages 1412--1421. Association for Computational
  Linguistics.

\bibitem[{Peters et~al.(2018)Peters, Neumann, Iyyer, Gardner, Clark, Lee, and
  Zettlemoyer}]{peters2018deep}
Matthew~E. Peters, Mark Neumann, Mohit Iyyer, Matt Gardner, Christopher Clark,
  Kenton Lee, and Luke Zettlemoyer. 2018.
\newblock Deep contextualized word representations.
\newblock In \emph{NAACL}.

\bibitem[{Radford et~al.(2019)Radford, Wu, Child, Luan, Amodei, and
  Sutskever}]{radford2019language}
Alec Radford, Jeffrey Wu, Rewon Child, David Luan, Dario Amodei, and Ilya
  Sutskever. 2019.
\newblock \href
  {https://d4mucfpksywv.cloudfront.net/better-language-models/language-models.pdf}
  {Language models are unsupervised multitask learners}.

\bibitem[{Snoek et~al.(2012)Snoek, Larochelle, and Adams}]{snoek2012practical}
Jasper Snoek, Hugo Larochelle, and Ryan~P Adams. 2012.
\newblock Practical bayesian optimization of machine learning algorithms.
\newblock In \emph{Advances in neural information processing systems}, pages
  2951--2959.

\bibitem[{So et~al.(2019)So, Liang, and Le}]{so2019evolved}
David~R. So, Chen Liang, and Quoc~V. Le. 2019.
\newblock \href {https://arxiv.org/abs/1901.11117} {The evolved transformer}.
\newblock In \emph{Proceedings of the 36th International Conference on Machine
  Learning (ICML)}.

\bibitem[{Strubell et~al.(2018)Strubell, Verga, Andor, Weiss, and
  McCallum}]{emnlp18-strubell}
Emma Strubell, Patrick Verga, Daniel Andor, David Weiss, and Andrew McCallum.
  2018.
\newblock { Linguistically-Informed Self-Attention for Semantic Role Labeling}.
\newblock In \emph{{Conference on Empirical Methods in Natural Language
  Processing (EMNLP)}}, Brussels, Belgium.

\bibitem[{Vaswani et~al.(2017)Vaswani, Shazeer, Parmar, Uszkoreit, Jones,
  Gomez, Kaiser, and Polosukhin}]{vaswani2017attention}
Ashish Vaswani, Noam Shazeer, Niki Parmar, Jakob Uszkoreit, Llion Jones,
  Aidan~N Gomez, Lukasz Kaiser, and Illia Polosukhin. 2017.
\newblock Attention is all you need.
\newblock In \emph{31st Conference on Neural Information Processing Systems
  (NIPS)}.

\end{thebibliography}
\bibliographystyle{acl_natbib}

\appendix


\end{document}